# A wearable input/output interface learns personalized models of haptic perception

Wen Li[1]*, E. Magnus Adams[2,3], Zijie Xu[4], Jiawen Yao[2], Aybala Esmer[2], Xiaoyue Ni[4,5,6]*, Bozhi Tian[1,7,8]*, Pedro Lopes[2,9]*

**Affiliations:**

[1]Department of Chemistry, The University of Chicago, Chicago, IL, 60637, USA.

[2]Department of Computer Science, The University of Chicago, Chicago, IL, 60637, USA.

[3]Department of Physics, The University of Chicago, Chicago, IL, 60637, USA.

[4]Department of Mechanical Engineering and Materials Science, Duke University, Durham, NC, 27708, USA.

[5]Department of Biostatistics and Bioinformatics, Duke University, Durham, NC, 27708, USA.

[6]Department of Electrical and Computer Engineering, Duke University, Durham, NC 27708, USA.

[7]The James Franck Institute, The University of Chicago, Chicago, IL, 60637, USA.

[8]The Institute for Biophysical Dynamics, The University of Chicago, Chicago, IL, 60637, USA.

[9]Neuroscience Institute, The University of Chicago, Chicago, IL, 60637, USA.

*Corresponding author. Email: pedrolopes@uchicago.edu, btian@uchicago.edu, xiaoyue.ni@duke.edu, wenli@uchicago.edu.

**Abstract:** Wearable haptic interfaces deliver physical stimuli to the skin, yet perceived locations vary widely across users and diverge from actuator locations, particularly for emergent tactile feedback techniques. We developed a wearable haptic input/output (I/O) device integrating a 36-channel electrotactile array with capacitive touch sensing, enabling stimulation and user-reported percept locations to be registered within the same body-referenced coordinate frame. In 24 participants, the device revealed electrotactile and referred thermal perceptual maps with notable interparticipant variability. Despite this variability, both modalities retained electrode-pair-specific spatial structure across participants. Tactile and thermal percepts also showed spatial correspondence, whose strength varied across participants. These findings motivated our human-in-the-loop calibration framework that adapts forward prediction strategy to individual users and generates personalized perceptual models. These models enabled inverse-rendering by selecting stimulation patterns predicted to generate a target percept for a specific user. Using these models, we demonstrate personalized haptic rendering in haptic-guided robot teleoperation, visual-to-haptic sensory substitution, interpersonal haptic communication, and virtual reality (VR). Together, this platform provides a route toward adaptive haptic interfaces that measure, model, and render according to each user's perception.

**One-Sentence Summary:** A wearable I/O interface registers and models electrotactile and thermal percepts for personalized haptic rendering.

## INTRODUCTION

Skin sensations are not merely a readout of physical stimulation. The same stimuli delivered to the same body site can evoke different percepts across users, varying in intensity, quality, location, and modality. This variability creates a central challenge for wearable haptic interfaces: a haptic device output cannot be assumed to produce the intended perceptual experience for every user. As haptic interfaces move toward spatially distributed stimulations with increasing complexity, they should answer an inverse question: given a desired sensation at a desired location, what stimulation pattern should be delivered to produce it for a particular user?

Addressing this problem requires linking stimulation technologies to the percepts they evoke. A wide range of technologies has been developed to generate haptic feedback on the skin, from mechanical actuators that render pressure, vibration, and shear to thermal actuators that deliver heating and cooling (*1*, *2*). Mechanical approaches include vibration actuators (*3–5*), pneumatic systems (*6*), ultrasonic displays (*7*), and hydraulic devices (*8*). As a non-mechanical alternative, electrotactile stimulation applies current to evoke tactile percepts through electrical activation of cutaneous sensory pathways (*9–15*), offering a route to thin, less cumbersome, low-power wearable devices (*11*, *12*). Thermal feedback, on the other hand, is typically delivered using thermoelectric devices (*14*, *16*, *17*), with emerging alternatives based on hydraulics (*18*) or chemical stimulation (*19*, *20*).

Despite these advances, wearable haptic systems are still primarily designed as stimuli-output platforms rather than as tools for measuring how users perceive stimulation. In many haptic systems, spatial cues are encoded by the position of the activated actuator (*11*, *14*, *18*, *21*), implicitly linking the intended perceptual location to the actuator layout—an approach we refer to as the actuator-centered. Although effective for many applications, this approach does not fully account for user-specific differences in perceived location (*22–24*). One approach to accommodating participant-specific differences is human-in-the-loop optimization, in which participant feedback enables adaptive adjustments or online learning of individual preferences (*25–27*). For spatial haptics, such feedback needs to capture where stimulation is perceived relative to where it is delivered. Mapping these percept locations commonly relies on verbal reports, manual pointing, or screen-based annotation (*5*, *10*, *12*, *14*, *15*, *22*, *28–33*), which are useful but can be sparse, offline, and not registered directly in a shared body-referenced coordinate frame.

This challenge is particularly pronounced for electrotactile stimulation. At the same body site, differences in skin impedance, skin thickness, receptor distribution, and peripheral nerve branching can change which sensory pathways are recruited (*10*, *12*, *23*, *34*). As a result, the same electrode-pair may evoke a localized tactile percept in one participant, a diffuse or displaced percept in another, and little or no percept in a third (*11*, *12*, *35*). Thus, even before adding other sensory channels, electrotactile stimulation requires user-specific characterization of where stimulation is perceived, which is typically achieved with the aid of an experimenter (*10*, *23*, *30*, *36–38*).

This key challenge is further exacerbated in multimodal haptics. Many systems combine tactile and thermal actuators to create richer multimodal haptic feedback (*14*, *18*, *39–41*). While they expand the range of sensations, the resulting percepts cannot necessarily be predicted from either modality alone (*28*, *37*, *42–44*). For example, in the temperature-texture illusion, reduced skin temperature diminishes roughness discrimination (*43*). And in the thermal-wetness illusion, cold and dry contact can be perceived as wet (*44*). Thermal referral extends these interactions to perceived location: when tactile stimulation is delivered near a thermal source, users can perceive thermal sensations at the site of the tactile percept rather than only at the physical location of the thermal actuator (*28*, *29*, *37*, *45–48*).

We recently showed that electrotactile can elicit thermal referral (*37*), shifting thermal sensations toward the perceived electrotactile location (Fig. 1B). This effect provides a route toward spatially distributed thermal feedback without covering large skin areas with dense arrays of bulky and power-intensive thermal actuators. However, deploying this requires lengthy experimenter-guided calibration to map the spatial relationship between electrotactile and thermal percepts across the stimulated area for each user (*10*, *23*, *30*, *36–38*).

Thus, we developed a haptic input/output (I/O) device that integrates a 36-channel electrotactile stimulation array with a capacitive touch layer. By registering stimulation and user-reported percepts in the same body-referenced coordinate frame, this I/O architecture enables human-in-the-loop perceptual mapping and modeling. Using this device, 24 participants mapped electrotactile and referred thermal percepts on their forearms by drawing directly on the capacitive touch layer. These self-reported perceptual maps enabled us to compare percept locations across participants. We found that both modalities showed substantial inter-participant variability but retained electrode-pair-specific spatial structure across participants. Tactile and thermal percepts also showed spatial correspondence, whose strength varied across participants, suggesting that tactile percept locations could help predict referred thermal percept locations in a participant-dependent way.

These findings motivated a perception-aware framework with two steps: forward prediction of percept locations from physical stimulation, and haptic inverse-rendering (i.e., selection of stimulation patterns for a desired percept). For forward prediction, we combined shared perceptual spatial structure with sparse participant-specific calibration. To predict tactile percept locations, we devised a global-offset-corrected population-mean model using only five tactile calibration trials. To predict referred thermal percept locations, we devised a Bayesian model also using only five thermal calibration trials. Additionally, we found this Bayesian model can be enhanced to predict more accurately if guided by tactile percept data. In effect, this demonstrates not only multimodal percept location prediction but also a pathway to reduce calibration time—a critical challenge in haptic interfaces (*36*, *49*, *50*)—as our approach utilizes sparse calibration trials (i.e., only a small subset of what a traditional exhaustive calibration would entail). We validated this with 12 new participants, finding that the thermal model derived from static perceptual maps better captured participant-specific thermal perceptual trajectories than actuator-centered predictions during sequential stimulation (i.e., thermal animations). Finally, we used these personalized models for inverse-rendering, translating desired percepts into stimulation patterns in four applications: personalized haptic rendering in VR, interpersonal haptic communication, visual-to-haptic sensory substitution, and haptic-guided robot teleoperation.

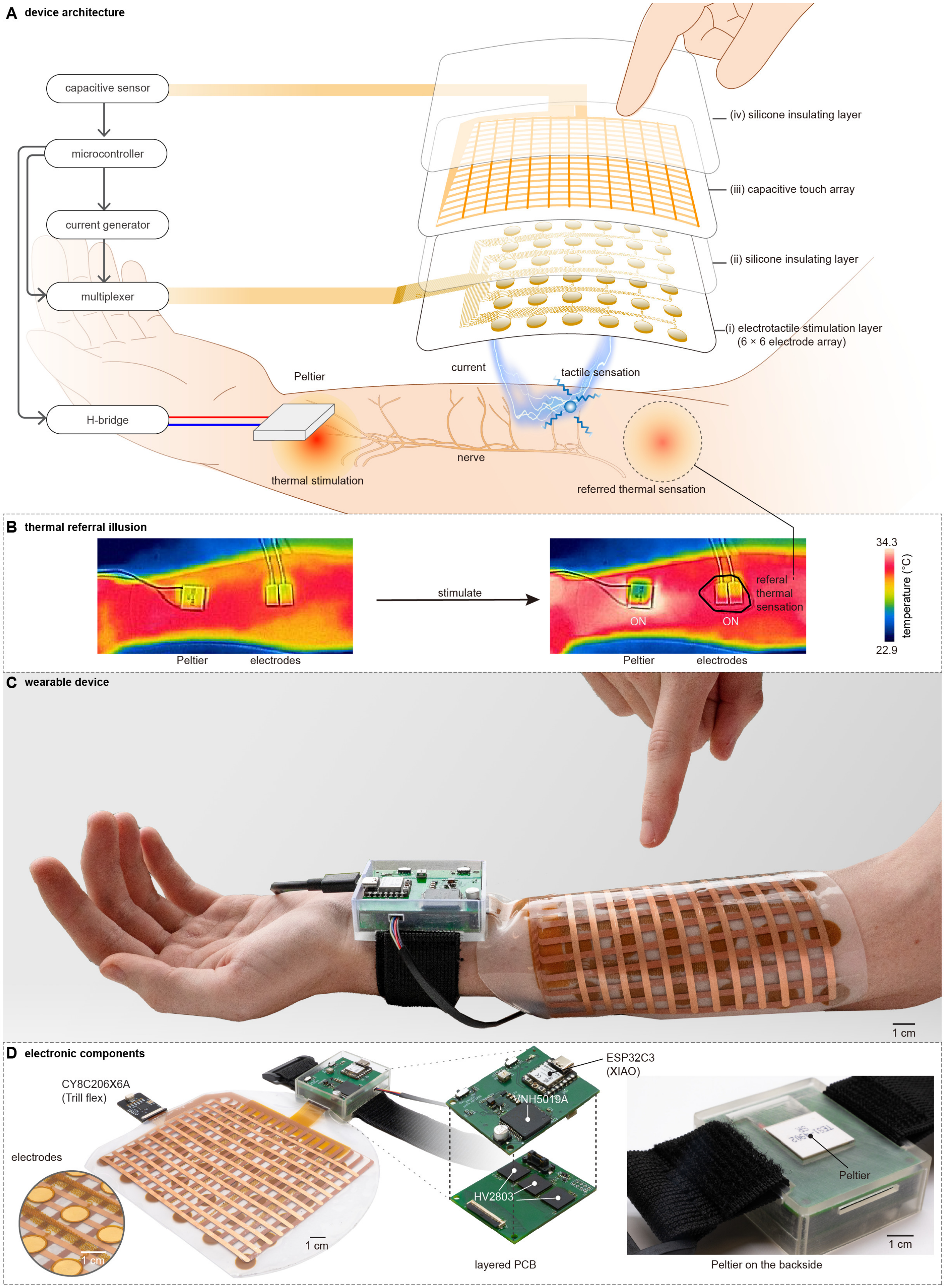

**Fig. 1. Overview of haptic I/O device architecture.** (**A**) Schematic of the multilayer wearable interface, integrating an electrotactile stimulation array with a capacitive touch array and external

control electronics. Electrotactile stimulation evokes localized tactile percepts, while the capacitive layer records user-reported percept locations. Combined with Peltier-based thermal stimulation, electrotactile stimulation can elicit thermal referral, shifting the perceived thermal location towards the tactile percept. (**B**) Thermal camera images of the forearm before and after the combined electrotactile-thermal stimulation. The measured temperature increase remained localized near the Peltier element, whereas the participant perceived warmth near the electrotactile percept location. (**C**) Photograph of a participant wearing the haptic I/O device on the forearm. (**D**) Wearable hardware architecture comprising the conformal haptic sleeve, two interconnected printed circuit boards (PCBs), and the Peltier element. The system is secured to the forearm using two adjustable hook-and-loop straps.

## RESULTS

### Device design and characterization

We designed a multilayer, conformal haptic sleeve integrating electrotactile output with capacitive touch input (Fig. 1A, 1C and fig. S1). The electrotactile layer comprised a 6 × 6 electrode array in a trapezoidal layout, with serpentine interconnects to enhance conformability (*51*). A 100 µm-thick silicone layer contained the array with 36 aligned openings for electrode-skin contact while insulating the interconnects. Above the array, a 500 µm-thick silicone layer isolated the high-voltage electrotactile circuitry from the capacitive touch layer. A 100 µm-thick silicone dielectric encapsulated the touch array, providing an electrically insulating surface for participants to report percept locations by touch. The key parameters of the haptic I/O device are shown in table S1.

The wearable control hardware comprised two interconnected PCBs containing electronics for electrotactile stimulation, high-voltage multiplexing, thermal control, capacitive touch sensing and wireless communication (Figs. 1D and 2A, fig. S2). Components for the PCBs are shown in Table S2. A high-voltage generator and current-mirror circuit produced current-limited, biphasic square-wave electrotactile stimulation, which three HV2803 multiplexers routed to the 36 electrode channels for programmable anode-cathode selection. A VNH5019A H-bridge circuit controlled the magnitude and direction of current through the Peltier element, enabling heating and cooling. The capacitive touch array was connected to a CY8C206X6A (Trill Flex) capacitive sensing IC, which transmitted the channel responses to an ESP32C3 (XIAO) microcontroller via $I^2C$. The microcontroller provided Bluetooth and Wi-Fi communication with external devices. The PCBs were enclosed in a custom 3D-printed housing, with the Peltier element mounted underneath for direct skin contact. Two adjustable hook-and-loop straps secured the device around the wrist (Fig. 1D). The device can be powered via other USB-devices (e.g., VR headsets) or power banks.

To reliably detect touch input during wear, we calibrated the capacitive array after device placement to account for the nonzero baseline caused by forearm contact. Following temporal filtering, baseline subtraction, and common-mode correction (see Supplementary Materials), touch onset was detected when a positive channel response exceeded three times its calibrated noise level.

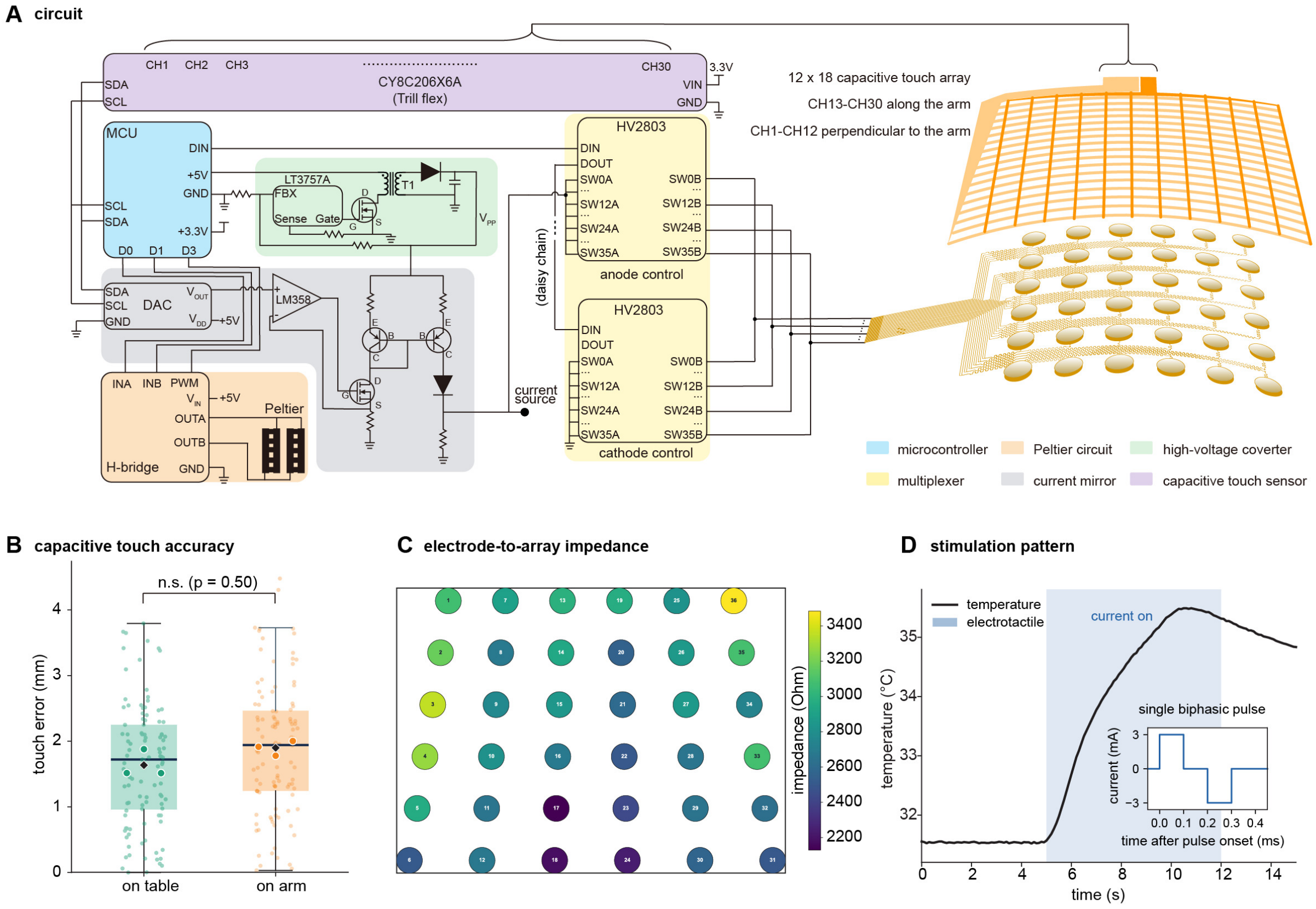

**Fig. 2. Characterization of the wearable haptic I/O system.** (**A**) Circuit schematic of the wearable control hardware for electrotactile stimulation, high-voltage multiplexing, thermal actuation, and capacitive touch sensing. (**B**) Spatial accuracy of the capacitive touch layer measured with the device placed on a flat surface and conformally mounted on the forearm. Touch localization error was calculated as the Euclidean distance between the center of the targeted electrode and the measured touch location (36 electrodes × 3 repeats). Small points represent individual touch errors. Large circles represent run means. Whiskers extend to 1.5 × IQR. No significant (n.s.) difference was observed between the two conditions (two-sided paired Wilcoxon signed-rank test, N = 3 paired runs, P = 0.50). (**C**) Spatial map of electrode-to-array impedance. For each measurement, the target electrode was connected to the positive terminal, while the remaining 35 electrodes served as a common negative return. Impedance was measured at 4.096 kHz using a 3.20 µA excitation current. (**D**) Combined electrotactile-thermal stimulation profile. Thermal stimulation was applied for 5 s, while electrotactile stimulation was delivered for 7 s, extending 2 s beyond the thermal stimulus. Electrotactile stimulation was delivered with 3 mA amplitude, 100 Hz and 100 µs pulse width.

We evaluated the spatial accuracy of the capacitive touch layer by sequentially touching the centers of all 36 electrodes three times each. Measurements were performed with the device placed on a flat table and conformally mounted on a forearm, yielding 108 touches per condition (Fig. 2B and fig. S3). All localization errors were below the 5 mm electrode radius—with maximum errors of 3.79 mm on the flat surface and 4.48 mm on the forearm. The mean localization errors were 1.64 mm and 1.90 mm, respectively, with no significant difference between conditions (P = 0.50). The electrode-to-array impedance ranged from 2 to 3.5 kΩ across the array (Fig. 2C), indicating consistent electrical contact (*52*). Peripheral electrodes showed slightly higher impedance than central electrodes, potentially reflecting site-dependent skin electrical properties as the array extended from the ventral toward the lateral and dorsal forearm (*52*). The sleeve exhibited stable force responses over 100 cycles of three-point bending, with channel resistance remaining stable before and after testing (fig. S4). Under electrotactile-only stimulation, we utilized a biphasic square-wave stimulation waveform with 3 mA amplitude, 100 Hz frequence and 100 μs pulse width; while under the combined electrotactile and thermal stimulation, the Peltier element heated the skin for 5 s, while electrotactile stimulation was delivered for 7 s (Fig. 2D), extending 2 s beyond heating to reduce ambiguity from the slower decay of thermal sensation (*28*, *37*).

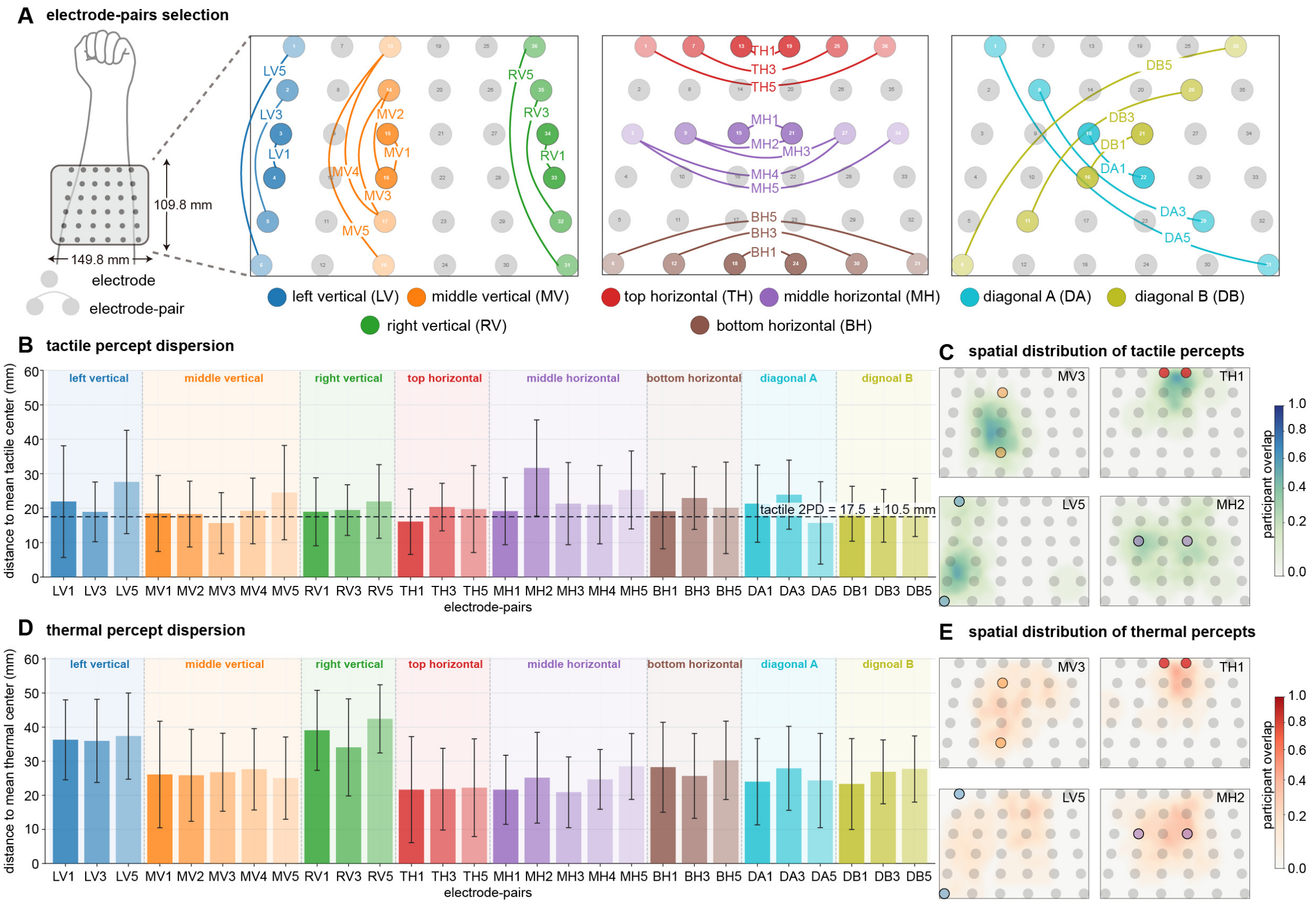

**Fig. 3. Electrode-pair selection and percept variability.** (**A**) Electrode-pair selection and pair identifiers (ID). (**B**) Dispersion of tactile percept locations for each electrode-pair (N = 24). The dashed reference indicates the reported forearm tactile two-point discrimination (2PD) distance. (**C**) Spatial distribution of tactile percepts for selected electrode-pairs. (**D**) Dispersion of thermal percept locations for each electrode-pair (N = 24). (**E**) Spatial distribution of thermal percepts for selected electrode-pairs. Bars in (B) and (D) are presented as mean ± SD.

## Inter-participant percept variability

We selected 28 representative electrode-pairs spanning vertical, horizontal, and diagonal orientations across the array (Fig. 3A). Pairs were grouped by their position and orientation as follows: left vertical (LV), middle vertical (MV), right vertical (RV), top horizontal (TH), middle horizontal (MH), bottom horizontal (BH), and two diagonal orientations (DA and DB). Additionally, a numerical suffix denotes the electrode-separation distance, yielding labels such as LV1 (i.e., denotes a vertical electrode-pair, positioned on the left side of the array, with an electrode-separation distance of 1) or MH5 (i.e., denotes a horizontal electrode-pair, positioned on middle of the array, with an electrode-separation distance of 5).

We evaluated electrotactile and referred thermal percepts in 24 participants (13 males and 11 females, mean age = 27.0 years, SD = 7.3 years). For standardized placement, the longitudinal centerline of the device was aligned with the midline of the ventral forearm, with the electrode array positioned 15 mm proximally from the elbow crease with the Peltier placed distally toward the wrist and 30 mm from the device edge (fig. S5A). Participants' forearm circumferences ranged from 19.5 to 31.3 cm (mean = 25.7 cm, SD = 3.1 cm—photographs are shown with participants' consent in fig. S5B).

We first established a thermal-only baseline: as expected, 19 of 24 participants reported thermal sensations around the Peltier element (which lied entirely outside the capacitive touch array), while 5 perceived thermal sensations in the Peltier area but also additional sensations within the capacitive touch array. Similar reports occurred in our previous work under the no-tactile condition (*37*). Pressure from the sleeve may have provided sufficient touch input to elicit some minor thermal referral in some participants.

Each participant experienced 28 electrode-pairs (randomized order) in two stimulation conditions (electrotactile-only and combined electrotactile-thermal), totaling 56 trials per participant. During electrotactile-only trials, participants received 2 s of stimulation and reported the tactile percept location. During combined electrotactile-thermal trials, participants received the 7 s stimulation pattern in Fig. 2D and reported the referred thermal percept location. Combined trials were followed by at least 30 s of recovery to allow skin temperature to return to baseline. Participants reported percepts by drawing freeform shapes with their dominant finger on the capacitive touch array over the non-dominant forearm or indicating when no percept occurred within the touch array. Stimulation was repeated after each report to confirm percept consistency, with further stimulation and reporting if the percept differed. This procedure required fewer than an average of 1.1 repeated presentations per trial under both conditions (fig. S6).

Most electrode-pairs produced a single connected percept region, whereas several horizontal and diagonal pairs with larger electrode-separation distance produced two separated regions (fig. S7), which were represented by two perceptual centroids in subsequent analyses. These pairs were TH3, TH5, MH3, MH4, MH5, BH3, BH5, DA3, DA5, DB3, and DB5. Stimulation reliably elicited tactile percepts (M = 98.5%, SD = 3.5%) and thermal percepts (M = 97%, SD = 11.2%) within the sensing area (fig. S8). Relative to the thermal-only baseline, referred thermals percepts were displaced by an average of 96.2 mm (fig. S9).

Despite this high response rate (≥ 97% for both conditions), percept locations varied substantially across participants and electrode-pairs. Interparticipant tactile percept distances ranged from 19.3 mm (P16-P21) to 44.8 mm (P11-P12) (fig. S10), with the maximum exceeding twice the reported mean forearm two-point discrimination threshold (17.5 ± 10.5 mm) (*53*). To characterize this variability for each electrode-pair, we measured the symmetric set distance between each participant's reported tactile percept and the group-mean tactile percept (Fig. 3B). Dispersion varied across electrode-pairs (fig. S11). Representative maps illustrate this contrast: MV3 and TH1 produced compact, consistently localized percepts, whereas LV5 and MH2 produced broader distributions (Fig. 3C).

Referred thermal percepts showed greater spatial variability, with interparticipant distances ranging from 20.9 mm (P5-P14) to 72.2 mm (P6-P17) (fig. S12). Distances from individual thermal percepts to the group-mean thermal percept were generally greater than those for tactile percepts (Fig. 3D). Variability was greatest for LV and RV pairs, where the device extended toward the lateral and dorsal forearm. Contrasting tactile to thermal maps also revealed spatial differences across modalities: thermal percepts were more diffuse and did not consistently coincide with their corresponding tactile percepts (fig. S11 and S13). Representative maps showed modest tactile-thermal displacement for MV3 and MH2 and larger displacement for LV5 (Fig. 3C and 3E).

Together, these results show that the same electrode-pairs evoke different tactile and thermal percept locations across participants, limiting predictions based solely on actuator location. This actuator-centered rendering is often used in prior studies attributing the percept location to either the physical location of the anode electrode, cathode electrode, or between the two electrodes (*54*–*57*). However, in our dataset, predictions based on these electrode locations each yielded mean localization errors exceeding 30 mm relative to user-reported percepts (fig. S14).

As referred thermal percepts emerge during combined electrotactile-thermal stimulation, we asked whether their perceived locations were spatially coupled to the corresponding electrotactile percepts. For each participant and electrode-pair, we defined the thermal-tactile distance as the symmetric set distance between the thermal and tactile percepts, with smaller values indicating stronger alignment. Thermal-tactile distance varied across electrode-pairs (Fig. 4A), with larger distances for peripheral LV and RV pairs and for increasing electrode-separation distances, particularly within the MV and MH groups. Conversely, closely spaced pairs near the central ventral forearm, such as MV1 and MH1, showed smaller distances.

Thermal-tactile alignment varied markedly across participants, with mean distances ranging from 9.2 mm for P21 to 62.5 mm for P5 (Fig. 4C). To test whether this alignment reflected electrode-pair-specific correspondence or a general spatial bias, we randomly permuted tactile electrode-pair labels within each participant (Fig. 4B) (*58*, *59*) (see Supplementary Materials). Matched thermal-tactile distances were smaller than the null distribution in 20 of 24 participants, with significant electrode-pair-specific correspondence in 19 participants after false-discovery-rate correction (Fig. 4C). At the population level, matched pairs showed a mean distance of 33.2 mm vs. 49.8 mm after random relabeling ($P = 0.0001$, Fig. 4D). Thus, referred thermal percepts indeed showed electrode-pair-specific spatial correspondence with tactile percepts, but the strength of correspondence varied across participants.

We next asked whether thermal and tactile perceptual maps have electrode-pair-specific spatial structure across participants. Across all 276 participant-pairs, we compared thermal percepts elicited by matched electrode-pairs with a null distribution generated by random electrode-pair relabeling (fig. S15A). Matched percepts were closer than the null distribution for 20 of 24 participants, with significant correspondence after false-discovery-rate correction (Fig. 4E). At the population level, matched pairs showed a mean distance of 41.8 mm vs. 45.8 mm after random relabeling ($P = 0.0001$, Fig. 4F). For tactile percepts, the same analysis showed even stronger cross-participant correspondence, with matched-pair distances 21.6 mm smaller than randomly relabeled distances ($P = 0.0001$, fig. S15).

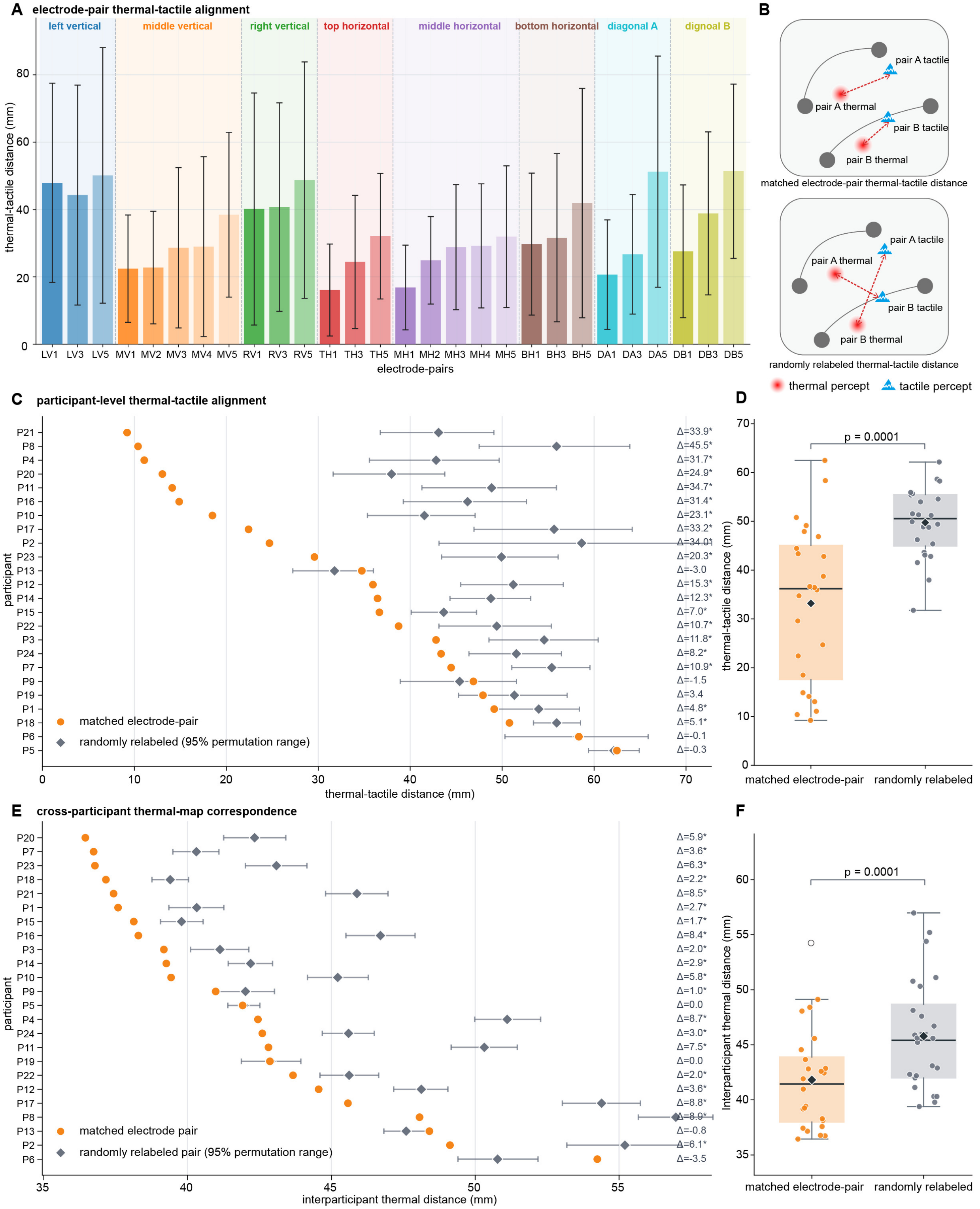

**Fig. 4. Thermal-tactile alignment and cross-participant spatial structure.** (**A**) Thermal-tactile distance across electrode-pairs. Bars show mean ± SD (N = 24). (**B**) Schematic of matched and randomly relabeled electrode-pair comparisons for thermal-tactile distance. (**C**) Participant-level thermal-tactile alignment. Matched thermal-tactile distances were compared with null distributions generated by permuting tactile electrode-pair labels within each participant. (**D**) Population-level thermal-tactile alignment. Matched distances were smaller than randomly relabeled distances (N

= 24, two-sided permutation test, P = 0.0001). (**E**) Cross-participant thermal map correspondence. Matched thermal-thermal distances were compared with null distributions generated by permuting thermal electrode-pair labels. (**F**) Population-level cross-participant thermal map correspondence. Matched distances were smaller than randomly relabeled distances (N = 24, two-sided permutation test, P = 0.0001). In (C) and (E), orange circles indicate matched distances, gray diamonds indicate permutation means, and error bars indicate central 95% permutation ranges; Δ denotes randomly relabeled distance minus matched distance, and asterisks indicate q < 0.05 after Benjamini–Hochberg false-discovery-rate correction. Permutation analyses used 10,000 permutations. In (D) and (F), points represent participant and participant-pair mean distances, respectively, and whiskers extend to 1.5 × IQR.

**Personalized perceptual models**

Tactile and thermal percepts displayed shared spatial patterns as well as individual differences. Therefore, we developed personalized models that combined population maps with a few calibration trials of a new prospective user. We evaluated these models in two directions: predicting the percept evoked by a given electrode-pair and selecting an electrode-pair to reproduce a target percept.

We compared three approaches to predict electrotactile percept locations: (i) the electrode-pair geometry (anode position, cathode position, and midpoint), (ii) the population-mean, and (iii) the population-mean adjusted by a participant-specific global offset estimated from five calibration trials (Fig. 5A). Under leave-one-participant-out (LOPO) evaluation, each participant was held out in turn. The midpoint approach used the geometric midpoint of the stimulating electrode-pair as the predicted percept location. The population-mean used the 23-participant average tactile percept locations as the predicted percept location. For the global-offset approach, five calibration trials were used to estimate a spatial offset between the population-mean predictions and the reported tactile locations. This offset was regularized and applied to predictions for the remaining electrode-pairs. Within each LOPO fold, the five calibration electrode-pairs and the regularization strength were selected by nested LOPO cross-validation (fig. S16). All three approaches were then evaluated on the held-out participant's uncalibrated electrode-pairs.

The population-mean approach significantly outperformed predictions based on the electrode-pair midpoint, anode, or cathode locations (P < 0.0001, Fig. 5B and fig. S17). Adjusting the population-mean by a participant-specific global offset modestly reduced the prediction error from 20.63 to 19.35 mm (P = 0.0016), improving predictions for 18 of 24 participants as well as improving predictions across electrode-pairs (fig. S18). Thus, shared perceptual structure provided a stronger basis for electrotactile prediction than actuator geometry.

We next asked whether these gains in forward prediction translated into more accurate inverse-rendering (i.e., selecting an electrode-pair to reproduce a target tactile percept). Using each measured percept from the 23 uncalibrated electrode-pairs as a target, each approach selected the electrode-pair whose predicted percept was closest to that target. The population-mean and the global offset approaches significantly outperformed the electrode-pair midpoint, reducing realized target error from 27.8 mm to 20.0 mm (P = 0.0009) and 18.4 mm (P = 0.0002), respectively, while increasing pair-selection agreement from 18.8% to 29.6% and 31.0%, respectively (fig. S19).

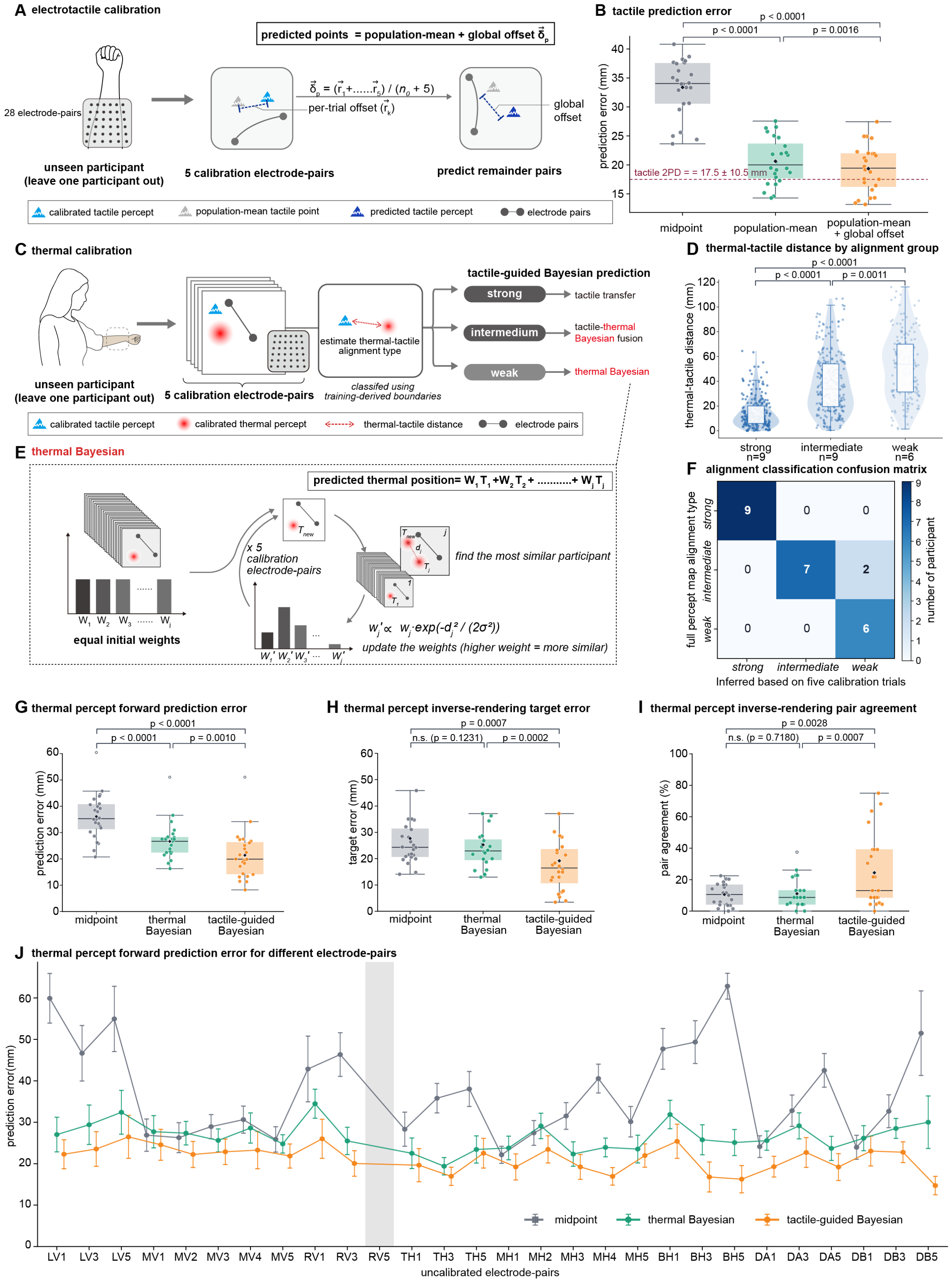

**Fig. 5. Calibration, prediction, and inverse-rendering of electrotactile and referred thermal percepts.** (**A**) Five-trial tactile calibration using a participant-specific global offset. Calibration

electrode-pairs and regularization weight $n_0$ were selected by inner LOPO validation. (**B**) Tactile prediction errors for the electrode-pair midpoint, population-mean, and offset-corrected population-mean. (**C**) Alignment classification from five calibration pairs' thermal-tactile distances using training-derived boundaries. (**D**) Thermal-tactile distance comparison among full-map alignment groups: strong (N = 9), intermediate (N = 9), and weak (N = 6). (**E**) Thermal Bayesian calibration. Starting from equal participant weights, each trial selected the electrode-pair with the largest posterior-weighted, thermal spatial variance and updated weights using the measured response. After five trials, posterior weights predicted uncalibrated percepts. (**F**) Five-trial vs. full-map alignment classification: 22/24 participants were correctly classified; two intermediate participants were classified as weak. LOPO-inferred groups determined tactile guidance. (**G**) Thermal forward prediction error. (**H**) Thermal inverse-rendering error using measured percepts from the 23 uncalibrated electrode-pairs as targets. Error was evaluated against the selected pair's measured percept. (**I**) Inverse-rendering pair agreement percentage. Values indicate the percentage of selecting the target-generating electrode-pair. (**J**) Thermal prediction error by electrode-pair (mean ± SEM). RV5 was always used for calibration and excluded from prediction evaluation. Panels (G–J) compare the electrode-pair midpoint, thermal Bayesian, and tactile-guided Bayesian methods. In (B), (G), (H), and (I), points represent participant-level results (N = 24). Whiskers in (B), (D), and (G–I) extend to 1.5 × IQR. Repeated-measures ANOVA with Greenhouse–Geisser correction: (B) $F(1.18, 27.24) = 165.98$, $P < 0.0001$, partial $\eta^2 = 0.878$; (G) $F(1.97, 45.35) = 50.77$, $P < 0.0001$, partial $\eta^2 = 0.688$; (H) $F(1.51, 34.74) = 14.33$, $P < 0.0001$, partial $\eta^2 = 0.384$; (I) $F(1.23, 28.19) = 13.92$, $P = 0.0004$, partial $\eta^2 = 0.377$. One-way between-participant ANOVA for (D): $F(2, 21) = 90.43$, $P < 0.0001$, partial $\eta^2 = 0.896$. Pairwise comparisons used two-sided paired t tests in (B) and (G–I), and two-sided Welch tests in (D), all with Holm correction.

Thermal prediction required a different strategy, because tactile percept locations carried information about referred thermal locations, but the strength of this relationship varied across participants. We therefore treated tactile percepts as an optional spatial cue: shared thermal-map structure supported tactile-independent prediction, and tactile information was incorporated only when informative. We tested whether thermal-tactile alignment could be inferred from five calibration trials and used to guide thermal prediction (Fig. 5C). One-dimensional k-means clustering (k = 3) of full-map alignment distances identified strong-, intermediate-, and weak-alignment groups comprising 9, 9, and 6 participants, respectively (Fig. 5D). Within each LOPO fold, the cluster centers and alignment boundaries were estimated exclusively from the training participants. For each held-out participant, thermal-tactile distances from five calibration electrode-pairs were corrected using training-derived pair biases and averaged into a pair-adjusted alignment distance, which was compared with the fold-specific boundaries to infer the participant's alignment group. The inferred alignment type matched the corresponding full-map classification for 22 of 24 held-out participants (Fig. 5F).

We first developed a thermal Bayesian calibration model that predicted referred thermal locations without tactile information, using shared thermal-map structure across participants (Fig. 5E). Under LOPO evaluation, the model initialized equal weights across training participants. At each calibration step, it selected the untested electrode-pair with the largest posterior-weighted spatial variance in the training participants' thermal percept, and then updated the weights according to the similarity between their thermal responses and the held-out participant's response. After five trials, the resulting posterior weights were used to predict thermal percept locations for the remaining electrode-pairs.

We then incorporated tactile information according to the inferred alignment group. For strong-alignment participants, the corresponding tactile percept directly predicted the thermal percept (tactile transfer). For intermediate-alignment participants, the thermal Bayesian prediction was shifted toward the corresponding tactile percept using a continuous electrode-pair-specific weight (tactile-thermal Bayesian fusion). Both strategies required a full tactile percept map, which is typically faster to acquire than a thermal map, as thermal requires longer stimulations, temperature recovery, and longer perceptual delays prior to a sensation arising (*60*, *61*). For weak-alignment participants, the thermal Bayesian prediction was retained without tactile guidance.

Without tactile maps, five-trial thermal Bayesian calibration achieved a mean prediction error of 26.65 mm and significantly outperformed predictions based on the electrode-pair midpoint, anode, or cathode location ($P < 0.0001$, Fig. 5G and fig. S20). With full tactile maps, the tactile-guided Bayesian framework further reduced the overall error to 21.41 mm, with improvements distributed across electrode-pairs (Fig. 5J). Group-specific prediction results were evaluated using the LOPO-inferred alignment groups. Relative to thermal Bayesian prediction, tactile transfer reduced error from 27.11 to 14.22 mm in strong-alignment participants ($P = 0.0010$, fig. S21), with improvements in all 9 participants. Tactile-thermal Bayesian fusion produced a smaller reduction in intermediate-alignment participants, from 23.76 to 22.37 mm ($P = 0.0350$), with improvements in 6 out of 7 participants.

We next tested whether these prediction gains translated into more accurate inverse-rendering. As targets, we took the thermal percepts generated by all 23 uncalibrated electrode-pairs. For both thermal Bayesian and tactile-guided thermal Bayesian, we calculated electrode-pair's target-to-percept distance in each predicted map and selected the pair with the smallest posterior-weighted mean distance. Without tactile guidance, the thermal Bayesian model provided no significant change in target error relative to the midpoint (25.3 vs. 27.7 mm, Fig. 5H). However, incorporating tactile guidance significantly reduced the error to 19.2 mm ($P = 0.0007$), with pair-selection agreement reaching 24.4% (Fig. 5H and 5I). The largest improvement was observed among strong-alignment participants, for whom both Bayesian approaches significantly outperformed the midpoint baseline, with tactile-guided rendering yielding the lowest error (8.7 mm, fig. S22).

Together, these analyses demonstrate that personalized perceptual maps supported both predicting where stimulation would be perceived and selecting electrode-pairs to render desired percept locations. Across tactile and thermal modalities, learned perceptual structure improved rendering beyond actuator geometry, with the largest thermal gains occurring when tactile and thermal percepts were strongly aligned.

**Sequential stimulation for perceptual animation**

Insofar, our model evaluated percepts evoked by individual electrode-pairs. Many haptic applications, however, require rendering spatial shapes, patterns, and motion rather than static percepts (*14*, *21*, *29*, *47*). We therefore tested whether models learned from static mapping could also capture participant-specific thermal perceptual trajectories during sequential stimulation.

We recruited an independent cohort of 12 participants (6 males, 6 females; mean age = 21.3 years, SD = 2.7 years) who were not included in the original mapping study. Their forearm circumferences ranged from 22.2 to 28.7 cm (mean = 24.9 cm, SD = 1.8 cm). Forearms photographs are shown with participants' consent in fig. S23. After completing the tactile-guided Bayesian calibration, participants experienced five combined electrotactile-thermal stimulation sequences (fig. S24), each repeated three times (15 trials total, presented in a randomized order). Each sequence activated four electrode-pairs sequentially, producing a spatially evolving referred thermal percept across the forearm (Fig. 6A).

After each trial, participants were shown two candidate reconstructions of the thermal animations in randomized order. One such animation reconstruction connected the thermal locations predicted by the tactile-guided Bayesian model, whereas the other animation reconstruction connected the geometric midpoints of the activated electrode-pairs. For calibration electrode-pairs, the measured location was used in both animation reconstructions to ensure the comparison reflected prediction performance only at uncalibrated sites (fig. S25). Participants selected the animation reconstruction that more closely matched their perception or selected "neither animation" if neither provided an adequate match. Model-predicted animations were selected significantly more often than midpoint-predicted animation or "neither animation" ($P = 0.0352$ and $P = 0.0015$, Fig. 6B). Thus, models learned from static mapping can be generalized from isolated percepts to animations produced by sequential stimulation, forming the basis for personalized spatiotemporal rendering.

**Demonstrated Applications**

Building on electrode-pair selection for individual target percepts and prediction of sequential percepts, we applied the personalized rendering framework in four proof-of-concept demonstrations. In these applications, information from Virtual Reality (VR), another person, a thermal camera, or a robot was represented as a target thermal percept location or trajectory and translated into a personalized electrode-pair stimulation sequence.

In this VR simulation (Fig. 6C), the calibration occurs in-context with the immersive interactions, i.e., as snowflakes fall and contact the user's forearm, users report tactile and thermal percept locations through the capacitive touch layer by flicking the snowflakes off their arm (following the in-VR instructions). These responses directly personalize the tactile-guided Bayesian model in real time, acting as calibration trials integrated within the VR simulation itself. This calibrated model is used for inverse-rendering within the same virtual environment, enabling more realistic subsequent interactions. In fact, as the user continues this simulation, they move from the outdoor environment to an indoor factory simulation where they can feel a thermal animation as they place a printed circuit board in a soldering oven—as this occurs, the system sequentially select electrode-pairs to render a thermal percept progressing along their forearm that aligns to the virtual motion.

This personalized rendering can be expanded to interpersonal haptic communication (Fig. 6D). The sender and receiver each wore a haptic I/O device connected through Wi-Fi. The sender drew a heart shape on the capacitive touch layer, and the recorded trajectory was transmitted to the receiver's device and resampled into six target thermal percept locations on the receiver's forearm. The receiver's personalized perceptual model mapped these targets to a sequence of electrode-pairs. These electrode-pairs were then activated sequentially to render the transmitted heart shape as a spatially evolving thermal percept.

Beyond human-generated input, our framework also can render sensor-derived spatial information (Fig. 6E). Existing visual-to-haptic systems, such as forehead electrotactile displays (*62*) or our back of the hand electrotactile display (*11*), translate camera-derived spatial information into electrotactile stimulation, enabling blind users to experience tactile images. However, displacement, diffusion, or merging of electrotactile percepts can distort these tactile images presented to the user. As such, our haptic I/O framework provides a perception-aware approach to sensory substitution by selecting electrode-pairs according to target percept locations. Here, a wrist-mounted thermal camera localizes a hot soldering iron, and the personalized perceptual model converts its location into a target thermal percept on the forearm and selects corresponding electrode-pairs. As the hand moves, the rendered thermal percept moves accordingly, providing a nonvisual spatial cue (fig. S26).

Moreover, these sensor-derived renderings can prove useful during robotic teleoperation. In a fire-rescue teleoperation demonstration, the operator's visual attention was largely occupied by the robot's forward-facing navigation camera (Fig. 6F). Four thermal cameras and LiDAR construct a robot-centered map of surrounding heat sources (fig. S27), whose directions are rendered on the operator's forearm. These cues provide directional information about heat sources outside the primary camera view while the operator visually monitors the forward-facing navigation camera.

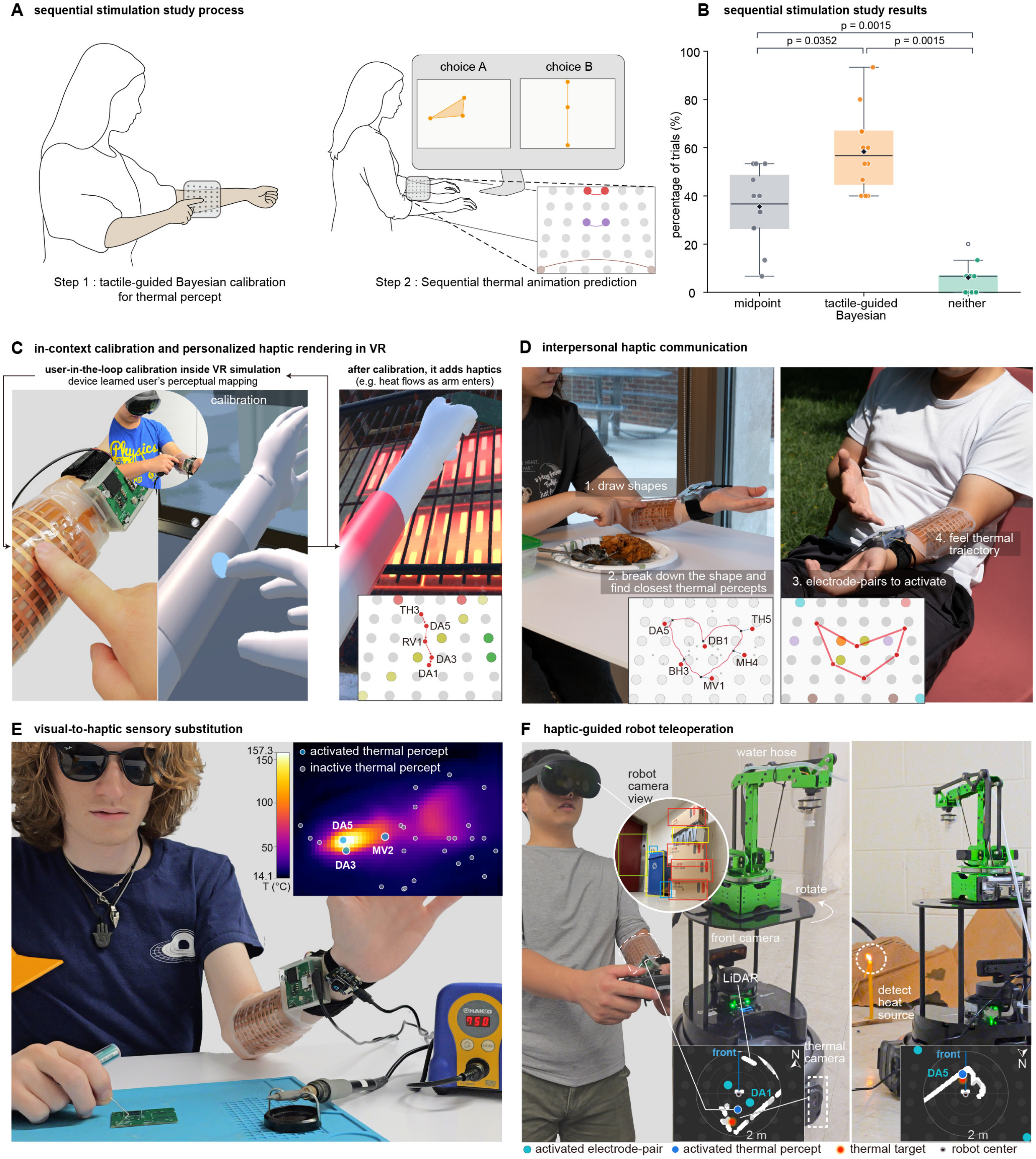

**Fig. 6. Sequential stimulation study and proof-of-concept applications.** (**A**) Sequential stimulation study. After tactile-guided Bayesian calibration, participants received combined electrotactile-thermal stimulation across four electrode-pairs in sequence. They then selected the thermal animation reconstruction that better matched their perception: model-predicted, electrode-pair-midpoint-predicted, or "neither animation". (**B**) Participant choices across sequential stimulation trials. Points represent each participant's percentage of trials assigned to each outcome ($N = 12$). Selection percentages differed across outcomes (Friedman test, $Q(2) = 19.45$, $P < 0.0001$, Kendall's $W = 0.810$). Pairwise comparisons used two-sided paired Wilcoxon signed-rank tests with Holm correction. Whiskers extend to $1.5 \times$ IQR. (**C**) In-context calibration and personalized

thermal rendering in VR. Users reported percept locations through the haptic I/O device during virtual interactions. After calibration, sequential electrotactile-thermal stimulation renders a spatially evolving heat percept along the forearm as the user moves a virtual PCB into a hot oven. (**D**) Interpersonal haptic communication. A sender drew a heart shape on the capacitive touch layer. Six target percept locations sampled from this trajectory were used with the receiver's personalized thermal map to select an electrode-pair sequence for thermal rendering. (**E**) Visual-to-haptic sensory substitution. A wrist-mounted thermal camera localized a hot soldering iron, and its position was translated into personalized thermal feedback on the forearm, providing a nonvisual spatial cue. (**F**) Haptic-guided robot teleoperation in a fire-rescue scenario. An operator controlled a mobile robot while viewing its forward-facing camera feed through a head-mounted display. Four thermal cameras and LiDAR localized surrounding heat sources, whose directions were rendered on the operator's forearm to convey thermal hazards outside the primary camera view.

## DISCUSSION

This work introduces a haptic I/O device that closes the loop between stimulation and perception by registering delivered stimuli and user-reported percept locations within a shared body-referenced coordinate frame. In 24 participants, electrotactile and referred thermal percepts retained electrode-pair-specific spatial structure despite substantial inter-participant variability. Combining this structure with sparse calibration improved both forward percept-location prediction and electrode-pair selection for inverse-rendering. Models learned from static maps also captured thermal animations during sequential stimulation. These capabilities supported four proof-of-concept applications translating information from VR, other people, and sensors into personalized stimulation sequences.

Tactile perceptual information improved thermal prediction, with benefits depending on individual thermal-tactile alignment. Thermal referral could also follow displaced tactile percepts, placing thermal sensations away from both the electrodes and the thermal actuator. These findings highlight the value of haptic I/O devices for characterizing user-specific multisensory interactions, with potential extensions to vibrotactile-thermal, nociceptive-tactile, and nociceptive-thermal coupling.

Finally, devices that co-register stimulation and percept locations could associate individual differences with peripheral nerve branching and the distribution of sensory nerve endings. Thermal sensation is mediated primarily by C and Aδ fibers (*63*), whereas discriminative touch relies predominantly on Aβ fibers (*64*). Differences in the spatial organization and recruitment of these pathways may contribute to the observed perceptual variability. Registering perceptual maps with high-resolution MRI of forearm nerves could test whether nerve branching patterns correlate with percept dispersion, and support the development of machine learning models that bridge anatomical features and individual perceptual maps.

## MATERIALS AND METHODS

### Haptic sleeve fabrication

Fabrication of the wearable haptic sleeve device began with spin-coating (WS-650-23; *Laurell*) a layer of Ecoflex (Ecoflex 00-30; Smooth-On; 1:1 w/w, 100 µm) onto a silicon wafer coated with poly(methyl methacrylate) (PMMA, 100 nm) as the sacrificial layer. The capacitive touch layer FPCB (ENIG-Cu, PET coverlay; *JLCPCB*; 240 µm) was laminated onto the Ecoflex after plasma activation (BD-20A; *Electro-Technic Products*) of the Ecoflex surface. A spin-coated Ecoflex insulation layer (500 μm) was then applied, onto which the 36-channel electrode array layer (ENIG-Cu, PI coverlay; *JLCPCB*; 110 μm) was aligned and bonded after plasma activation of the insulation surface. An Ecoflex encapsulation layer (100 μm) was spin-coated. Windows over the electrotactile pads and connector pins of both conductive layers were then opened using a picosecond laser (ProtoLaser R4; *LPKF*), exposing the pads through the Ecoflex for skin contact and electrical interfacing. The completed device was released from the silicon wafer by undercutting the PMMA sacrificial layer in acetone. Detailed step-by-step fabrication procedures are provided in the Supplementary Materials.

**Wearable stimulation and sensing electronics**

The wearable electronics comprised two interconnected PCBs interfaced with the conformal haptic I/O device and controlled by a ESP32C3 microcontroller (XIAO, Seeed Studio). Electrotactile stimulation was supplied either by an internal custom constant-current stimulator or an external stimulator. The internal stimulator used an LT3757A boost converter to generate a high-voltage rail and an LM358 feedback circuit with a high-voltage MOSFET to regulate stimulation current. A PNP current mirror implemented with FCX705TA transistors replicated the reference current at the output. Three 32-channel HV2803 high-voltage analog switches routed stimulation to the 36 electrodes and enabled programmable anode–cathode selection. For human-subject user studies, we instead used an external computer-controlled constant-current stimulator (RehaStim, Hasomed) in accordance with the approved experimental protocol. Both stimulation modes supported programmable current, pulse width, and frequency.

Thermal stimulation was delivered using a 2 × 2 cm Peltier element (TES1-4902) driven by a VNH5019A H-bridge (Pololu Robotics and Electronics). The microcontroller controlled current direction and duty cycle to control heating and cooling. A heat sink was added during human-subject experiments to accelerate return to baseline temperature between trials.

Capacitive touch sensing was implemented using a Trill Flex evaluation board (Bela) based on the CY8C206x6A capacitive sensing IC (Infineon), providing up to 30 sensing channels with programmable gain and thresholds. The microcontroller communicated with the touch sensor through $I^2C$ and controlled electrode-pair selection, stimulation polarity, intensity, pulse width, timing, and thermal stimulation. Touch data were transmitted via USB during human-subject experiments and via Bluetooth or Wi-Fi in application demonstrations. The system also supported wireless communication with external devices through Bluetooth, Wi-Fi, with Open Sound Control used for network messaging.

**Psychophysics user study procedure**

The experimental protocol (IRB22-0228) was approved by the Institutional Review Board at the University of Chicago. We recruited 24 healthy volunteers, including 13 male and 11 female participants, between 21 and 58 years of age (mean age = 27.0 years, SD = 7.3 years). Participants were compensated with $20 for their time. Ethnicity information was not collected. Written informed consent was obtained from all participants before the study. Participants had no visible cuts, wounds, or self-reported skin conditions on the tested forearms.

With consent, we photographed each participant's non-dominant forearm and measured its circumference near the elbow (19.5–31.3 cm; mean = 25.7cm, SD = 3.1 cm; fig. S5). The skin was cleaned with alcohol, followed by a thin layer of ultrasound gel (Aquasonic) to improve electrode-skin contact. The haptic I/O device was positioned on the ventral non-dominant forearm, with its longitudinal centerline aligned to the forearm midline and its proximal edge 15 mm from the elbow crease and secured with medical tape. The Peltier element was then attached, and heating was briefly tested to confirm participant comfort.

The study was controlled using a custom Python graphical user interface. Before each session, a no-touch baseline was recorded for capacitive sensing. Participants completed two practice trials, one electrotactile-only and one combined electrotactile-thermal. They then received thermal-only stimulation and either drew the perceived thermal region on the capacitive touch layer or indicated that it fell outside the sensing area.

Each participant tested 28 electrode-pairs in randomized order. For each electrode-pair, an electrotactile-only trial was followed by a combined electrotactile-thermal trial before proceeding to the next electrode-pair. Electrotactile-only stimulation was delivered for 2 s at 3 mA, 100 Hz, and 100 μs pulse width using biphasic current stimulation. During combined stimulation, electrotactile and thermal stimulation lasted 7 and 5 s, respectively. The Peltier element was driven at 8 V, with a PWM duty cycle of 100% for the first 1 s and 31.4% for the remaining 4 s. The additional 2 s of electrotactile stimulation reduced ambiguity from the slower decay of thermal sensation. After each combined trial, participants waited at least 30 s to allow skin temperature to return toward baseline.

Participants reported tactile percept boundaries during electrotactile-only trials and thermal percept boundaries during combined trials by drawing on the capacitive touch layer. They could skip trials when no sensation was perceived within the sensing area. After each drawing, stimulation was repeated, and participants selected “Yes” or “No” in the graphical interface to indicate whether the percept matched the reported location. If “No” was selected, the response was cleared, stimulation was delivered again, and the participant redrew the percept. This procedure continued until the participant selected “Yes,” and the confirmed response was used for analysis.

**Data analysis and statistics**

Electrotactile and thermal percept locations were transformed into a common two-dimensional forearm coordinate system. Each reported percept point or area was treated as one percept, with areas represented by their polygon centroids. Each trial contained one or two percept centroids, which together constituted the percept set for that trial.

For each electrode-pair, we calculated the two-percept response rate as the proportion of tactile responses containing two spatially separated percepts. Eleven horizontal or diagonal electrode-pairs with rates > 40% (TH3, TH5, MH3, MH4, MH5, BH3, BH5, DA3, DA5, DB3, and DB5) were represented by two population centroids in subsequent analyses. For these pairs, percept centroids were pooled across participants and separated using two-dimensional Euclidean k-means clustering ($k = 2$), with the cluster means defining the population centroids. All other electrode-pairs were represented by a single population centroid. The resulting one- or two-centroid representation defined the population percept set for each electrode-pair. The same structure was applied to thermal percepts to maintain equivalent model complexity across modalities, with thermal percepts for the same 11 electrode-pairs independently clustered using $k = 2$.

Distances between percept sets were quantified using a symmetric nearest-neighbor set distance. For two percept sets A and B, we calculated the mean distance from each centroid in A to its nearest centroid in B, repeated the calculation from B to A, and averaged the two directional values:

$$D(A,B) = \frac{1}{2}[\frac{1}{|A|}\sum_{a\in A}\min_{b\in B}\|a-b\| + \frac{1}{|B|}\sum_{b\in B}\min_{a\in A}\|b-a\|]$$

Percept dispersion in Fig. 3B and 3D was quantified as the symmetric set distance between each participant's tactile or thermal percept set and the corresponding population-mean percept set for the same electrode-pair. Thermal-tactile distance in Figs. 4A, 4C, 4D, and 5D was calculated between the tactile and thermal percept sets. Cross-participant thermal/tactile distance in Figs. 4E, 4F and fig. S15 were calculated between participants' thermal/tactile percept sets.

For repeated-measures ANOVAs, Greenhouse–Geisser-corrected degrees of freedom and P values were reported. Statistical tests and multiple-comparison corrections are specified in the corresponding figure captions. Statistical significance was evaluated at $P < 0.05$, using Holm-adjusted P values where applicable, or $q < 0.05$ for false-discovery-rate-corrected analyses.

**Electrotactile percept prediction**

Electrotactile percept prediction was evaluated using LOPO cross-validation. In each fold, the population tactile map was constructed from the remaining 23 participants, using the fixed one- or two-centroid structure defined above. We compared the electrode-pair midpoint, the population-mean tactile map, and the population map with a participant-specific global offset.

Five calibration electrode-pairs were selected by greedy, nested LOPO validation within the training library. For each calibration electrode-pair $p$ and test participant $i$, observed and population centroids were matched, and their displacements were averaged to obtain one pair-level displacement vector, $r(i, p)$. The global offset was regularized toward zero as:

$$\vec{\delta}_p = \frac{\sum_{p\in C_i}\vec{r}(i,\ p)}{|C_i|\ +\ n_0}$$

where $C_i$ is the set of calibration electrode-pairs with valid percept locations. A regularization weight of $n_0 = 4$ was used in the final analysis. Inner LOPO analysis showed that $n_0 = 4$ minimized the mean inner-validation error (Fig. S16). The resulting global offset was added to every centroid in the population-mean tactile map to generate the participant-specific prediction.

Prediction error was calculated as the mean distance from each observed percept centroid set $O$ to its nearest predicted centroid set $X$ to quantify how closely the model localized each experienced percept set:

$$Prediction\ error = \frac{1}{|O|}\sum_{o\in O}\min_{x\in X}\|o-x\|$$

Calibration electrode-pairs were excluded from evaluation, and errors were averaged first across uncalibrated electrode-pairs within each participant and then across participants.

**Thermal Bayesian model**

Thermal prediction was evaluated using the same LOPO framework. In each fold, the 23 training participants were initialized with equal weights. After each thermal calibration trial, participant weights were updated:

$$w_j^{(k)} \propto w_j^{(k-1)} \exp\left[-\frac{D(T_{i,p_k}, T_{j,p_k})^2}{2\sigma^2}\right]$$

where $i$ = test participant, $j$ = training participant, $p_k$ = $k_{th}$ calibration electrode-pair and σ = 25 mm. Weights were normalized after each update. After five calibration-trials, the posterior-weighted training responses were used to predict thermal percept locations for the unrevealed electrode-pairs. The same five calibration pairs were used for the tactile-guided Bayesian model. Calibration selection is described in the Supplementary Materials. Thermal prediction error was calculated using the same observed to nearest predicted centroid metric described for electrotactile prediction.

**Tactile-guided Bayesian method**

Tactile guidance was determined from thermal-tactile alignment at the five electrode-pairs selected during thermal calibration, using their thermal responses and corresponding measured tactile percepts. For each valid calibration electrode-pair, the thermal-tactile distance was corrected for an electrode-pair-specific bias estimated from the training participants. The participant's pair-adjusted alignment distance was:

$$AD_i = \frac{1}{|C_i|} \sum_{p \in C_i} [D(S_{i,p}, T_{i,p}) - b_p]$$

where $D(S_{i,p}, T_{i,p})$ is the thermal-tactile distance between tactile percept $S_{i,p}$ and thermal percept $T_{i,p}$, $b_p$ is the training-derived pair bias (see Supplementary Materials), and $C_i$ is the set of calibration electrode-pairs with valid percept locations. Training participants' full map thermal-tactile distance were clustered using one-dimensional k-means (k = 3), with $D_1$ and $D_2$ defined as the midpoints between adjacent k-means cluster centers. The pair-adjusted alignment distance $AD_i$ was compared with $D_1$ and $D_2$ to determine the test participant's alignment type. Group-specific prediction results were evaluated using these LOPO-inferred groups.

The participant-specific tactile percept map measured across electrode-pairs was used for tactile guidance. For strong-alignment participants, the corresponding tactile percept set was used directly to predict thermal location. For weak-alignment participants, the thermal Bayesian prediction was retained. For intermediate-alignment participants, the Bayesian prediction was shifted toward the corresponding tactile percept according to a continuous pair-specific weight:

$$\lambda_p = 0.5e^{-d_p^2/(2\sigma^2)}$$

where $d_p$ is the mean thermal-tactile distance for electrode-pair $p$ among intermediate-alignment training participants and σ = 25 mm. The tactile-guided prediction $\hat{T}_{ip}^{TG}$ was calculated as the fusion of thermal Bayesian prediction and tactile percept location:

$$\hat{T}_{i,p}^{TG} = (1 - \lambda_p)\hat{T}_{i,p}^{B} + \lambda_p S_{i,p}$$

where $\hat{T}_{i,p}^{B}$ is the Bayesian thermal prediction. For multi-centroid percept sets, Bayesian and tactile centroids were matched before the weighted fusion was applied.

## Supplementary Materials

Materials and Methods

Table 1-2

Figs. S1 to S27

## References

1. H. Culbertson, S. B. Schorr, A. M. Okamura, Haptics: The Present and Future of Artificial Touch Sensation. *Annual Review of Control, Robotics, and Autonomous Systems* **1**, 385–409 (2018).

2. S. Patel, Z. Rao, M. Yang, C. Yu, Wearable Haptic Feedback Interfaces for Augmenting Human Touch. *Adv. Funct. Mater.* **36**, 2417906 (2026).

3. J. Rekimoto, “Traxion: a tactile interaction device with virtual force sensation” in *Proceedings of the 26th Annual ACM Symposium on User Interface Software and Technology* (Association for Computing Machinery, New York, NY, USA, 2013; https://dl.acm.org/doi/10.1145/2501988.2502044)*UIST ’13*, pp. 427–432.

4. X. Yu, Z. Xie, Y. Yu, J. Lee, A. Vazquez-Guardado, H. Luan, J. Ruban, X. Ning, A. Akhtar, D. Li, B. Ji, Y. Liu, R. Sun, J. Cao, Q. Huo, Y. Zhong, C. Lee, S. Kim, P. Gutruf, C. Zhang, Y. Xue, Q. Guo, A. Chempakasseril, P. Tian, W. Lu, J. Jeong, Y. Yu, J. Cornman, C. Tan, B. Kim, K. Lee, X. Feng, Y. Huang, J. A. Rogers, Skin-integrated wireless haptic interfaces for virtual and augmented reality. *Nature* **575**, 473–479 (2019).

5. Y. H. Jung, J.-Y. Yoo, A. Vázquez-Guardado, J.-H. Kim, J.-T. Kim, H. Luan, M. Park, J. Lim, H.-S. Shin, C.-J. Su, R. Schloen, J. Trueb, R. Avila, J.-K. Chang, D. S. Yang, Y. Park, H. Ryu, H.-J. Yoon, G. Lee, H. Jeong, J. U. Kim, A. Akhtar, J. Cornman, T. Kim, Y. Huang, J. A. Rogers, A wireless haptic interface for programmable patterns of touch across large areas of the skin. *Nat. Electron.* **5**, 374–385 (2022).

6. C. du Pasquier, L. Tessmer, I. Scholl, L. Tilton, T. Chen, S. Tibbits, A. Okamura, Haptiknit: Distributed stiffness knitting for wearable haptics. *Sci. Robot.* **9**, eado3887 (2024).

7. C. Park, Y. Lee, S. H. Yoon, UltraBoard: Always-available Wearable Ultrasonic Mid-air Haptic Interface for Responsive and Robust VR Inputs. *Proc. ACM Interact. Mob. Wearable Ubiquitous Technol.* **9**, 44:1-44:31 (2025).

8. J. Nash, K. Sauvé, C. M. van Riet, A. van Oosterhout, A. Sharma, C. Clarke, J. Alexander, “HydroHaptics: High-Fidelity Force-Feedback on Soft Deformable Interfaces using Hydrostatic Transmission” in *Proceedings of the 38th Annual ACM Symposium on User Interface Software and Technology* (Association for Computing Machinery, New York, NY, USA, 2025; https://dl.acm.org/doi/10.1145/3746059.3747679)*UIST ’25*, pp. 1–20.

9. R. Blau, A. Abdal, N. Root, A. X. Chen, T. Rafeedi, R. Ramji, Y. Qie, T. Kim, A. Navarro, J. Chin, L. L. Becerra, S. J. Edmunds, S. M. Russman, S. A. Dayeh, D. P. Fenning, R. Rouw, D. J. Lipomi, Conductive block copolymer elastomers and psychophysical thresholding for accurate haptic effects. *Sci. Robot.* **9**, eadk3925 (2024).

10. H. Kajimoto, Electrotactile Display with Real-Time Impedance Feedback Using Pulse Width Modulation. *IEEE Transactions on Haptics* **5**, 184–188 (2012).

11. S.-Y. Teng, G. S.-H. Kim, X. Liu, P. Lopes, "Seeing with the Hands: A Sensory Substitution That Supports Manual Interactions" in *Proceedings of the 2025 CHI Conference on Human Factors in Computing Systems* (ACM, Yokohama Japan, 2025; https://dl.acm.org/doi/10.1145/3706598.3713419), pp. 1–15.

12. Y. Tanaka, A. Shen, A. Kong, P. Lopes, "Full-hand Electro-Tactile Feedback without Obstructing Palmar Side of Hand" in *Proceedings of the 2023 CHI Conference on Human Factors in Computing Systems* (ACM, Hamburg Germany, 2023; https://dl.acm.org/doi/10.1145/3544548.3581382), pp. 1–15.

13. Y. Shi, F. Wang, J. Tian, S. Li, E. Fu, J. Nie, R. Lei, Y. Ding, X. Chen, Z. L. Wang, Self-powered electro-tactile system for virtual tactile experiences. *Sci. Adv.* **7**, eabe2943 (2021).

14. Y. Huang, J. Zhou, P. Ke, X. Guo, C. K. Yiu, K. Yao, S. Cai, D. Li, Y. Zhou, J. Li, T. H. Wong, Y. Liu, L. Li, Y. Gao, X. Huang, H. Li, J. Li, B. Zhang, Z. Chen, H. Zheng, X. Yang, H. Gao, Z. Zhao, X. Guo, E. Song, H. Wu, Z. Wang, Z. Xie, K. Zhu, X. Yu, A skin-integrated multimodal haptic interface for immersive tactile feedback. *Nat. Electron.* **6**, 1020–1031 (2023).

15. K. Yao, J. Zhou, Q. Huang, M. Wu, C. K. Yiu, J. Li, X. Huang, D. Li, J. Su, S. Hou, Y. Liu, Y. Huang, Z. Tian, J. Li, H. Li, R. Shi, B. Zhang, J. Zhu, T. H. Wong, H. Jia, Z. Gao, Y. Gao, Y. Zhou, W. Park, E. Song, M. Han, H. Zhang, J. Yu, L. Wang, W. J. Li, X. Yu, Encoding of tactile information in hand via skin-integrated wireless haptic interface. *Nat. Mach. Intell.* **4**, 893–903 (2022).

16. S.-W. Kim, S. H. Kim, C. S. Kim, K. Yi, J.-S. Kim, B. J. Cho, Y. Cha, Thermal display glove for interacting with virtual reality. *Sci. Rep.* **10**, 11403 (2020).

17. H. Gao, S. Cai, Y. Wu, K. Zhu, "ThermOuch: A Wearable Thermo-Haptic Device for Inducing Pain Sensation in Virtual Reality through Thermal Grill Illusion" in *SIGGRAPH Asia 2024 Conference Papers* (Association for Computing Machinery, New York, NY, USA, 2024; https://dl.acm.org/doi/10.1145/3680528.3687620)*SA '24*, pp. 1–12.

18. J. Jeong, M. Kang, Y. Song, S. Pyo, Interactive Haptic System with Multimodal Tactile Sensing and Hydraulic Feedback for Realistic Human–Machine Interaction. *Adv. Intell. Syst.* **7**, 2400578 (2025).

19. J. Lu, Z. Liu, J. Brooks, P. Lopes, "Chemical Haptics: Rendering Haptic Sensations via Topical Stimulants" in *The 34th Annual ACM Symposium on User Interface Software and Technology* (ACM, Virtual Event USA, 2021; https://dl.acm.org/doi/10.1145/3472749.3474747), pp. 239–257.

20. J. Brooks, S. Nagels, P. Lopes, "Trigeminal-based Temperature Illusions" in *Proceedings of the 2020 CHI Conference on Human Factors in Computing Systems* (ACM, Honolulu HI USA, 2020; https://dl.acm.org/doi/10.1145/3313831.3376806), pp. 1–12.

21. M. T. Flavin, K.-H. Ha, Z. Guo, S. Li, J.-T. Kim, T. Saxena, D. Simatos, F. Al-Najjar, Y. Mao, S. Bandapalli, C. Fan, D. Bai, Z. Zhang, Y. Zhang, E. Flavin, K. E. Madsen, Y. Huang, L. Emu, J. Zhao, J.-Y. Yoo, M. Park, J. Shin, A. G. Huang, H.-S. Shin, J. E. Colgate, Y. Huang, Z. Xie, H. Jiang, J. A. Rogers, Bioelastic state recovery for haptic sensory substitution. *Nature* **635**, 345–352 (2024).

22. F. Iberite, J. Muheim, O. Akouissi, S. Gallo, G. Rognini, F. Morosato, A. Clerc, M. Kalff, E. Gruppioni, S. Micera, S. Shokur, Restoration of natural thermal sensation in upper-limb amputees. *Science* **380**, 731–735 (2023).

23. A. Akhtar, J. Sombeck, B. Boyce, T. Bretl, Controlling sensation intensity for electrotactile stimulation in human-machine interfaces. *Sci. Robot.* **3**, eaap9770 (2018).

24. M. Mikkelsen, J. He, M. Tommerdahl, R. A. E. Edden, S. H. Mostofsky, N. A. J. Puts, Reproducibility of flutter-range vibrotactile detection and discrimination thresholds. *Sci. Rep.* **10**, 6528 (2020).

25. R. Zhang, X. Zhu, M. Pourebadi Khotbehsara, W. Dao, E. Bıyık, H. Culbertson, "Vibrotactile Preference Learning: Uncertainty-Aware Preference Learning for Personalized Vibration Feedback" in *Proceedings of the 34th ACM Conference on User Modeling, Adaptation and Personalization* (Association for Computing Machinery, New York, NY, USA, 2026; https://dl.acm.org/doi/10.1145/3774935.3806784)*UMAP '26*, pp. 445–449.

26. H. Tolasa, B. Catkin, V. Patoglu, Human-in-the-Loop Optimization of Perceived Realism of Multi-Modal Haptic Rendering Under Conflicting Sensory Cues. *IEEE Transactions on Haptics* **18**, 295–311 (2025).

27. B. Catkin, V. Patoglu, Preference-Based Human-in-the-Loop Optimization for Perceived Realism of Haptic Rendering. *IEEE Transactions on Haptics* **16**, 470–476 (2023).

28. H. Wang, Y. Singhal, H. Gil, J. R. Kim, "Thermal Masking: When the Illusion Takes Over the Real" in *Proceedings of the CHI Conference on Human Factors in Computing Systems* (ACM, Honolulu HI USA, 2024; https://dl.acm.org/doi/10.1145/3613904.3641941), pp. 1–16.

29. Y. Singhal, D. Honrales, J. R. Kim, "HeatFlow: A Thermal-Tactile Display for Dynamic 2D Thermal Movements" in *Proceedings of the 38th Annual ACM Symposium on User Interface Software and Technology* (Association for Computing Machinery, New York, NY, USA, 2025; https://dl.acm.org/doi/10.1145/3746059.3747619)*UIST '25*, pp. 1–18.

30. K. Ushiyama, P. Lopes, "FeetThrough: Electrotactile Foot Interface that Preserves Real-World Sensations" in *Proceedings of the 36th Annual ACM Symposium on User Interface Software and Technology* (Association for Computing Machinery, New York, NY, USA, 2023; https://dl.acm.org/doi/10.1145/3586183.3606808)*UIST '23*, pp. 1–11.

31. B. Shan, Y. Guo, Y. Wang, P. Zhao, Y. Wang, Z. Wang, L. He, Y. Liu, Y. Wang, W. Guo, Y. Zhang, Z. Dai, X. Yu, D. Wang, Fully Transparent Haptic Interface for High-Resolution Tactile Feedback on Touchscreens. *Adv. Sci.* **12**, e11874 (2025).

32. H. I. Yi, H. Lee, S. H. Yoon, "ThermicVib: Enabling Dynamic Thermal Sensation with Multimodal Haptic Glove for Thermal-Responsive Interaction" in *2024 IEEE International Symposium on Mixed and Augmented Reality (ISMAR)* (2024; https://ieeexplore.ieee.org/document/10765498/), pp. 614–623.

33. Q. Ouyang, J. Wu, S. Sun, J. Pensa, A. Abiri, E. P. Dutson, J. W. Bisley, Bio-inspired Haptic Feedback for Artificial Palpation in Robotic Surgery. *IEEE Trans. Biomed. Eng.* **68**, 3184–3193 (2021).

34. H. Kajimoto, N. Kawakami, S. Tachi, "Electro-Tactile Display with Tactile Primary Color Approach" in *Workshop at the IEEE/RSJ International Conference on Intelligent Robots and Systems (IROS)* (2004).

35. J. Ara, S. H. Hwang, T. Song, G. Khang, Electrically-elicited tactile sensation for different modulation types, polarities and waveforms of stimulation pulse trains. *Int. J. Precis. Eng. Manuf.* **13**, 1911–1916 (2012).

36. S. Vizcay, P. Kourtesis, F. Argelaguet, C. Pacchierotti, M. Marchal, Design, evaluation and calibration of wearable electrotactile interfaces for enhancing contact information in virtual reality. *Computers & Graphics* **111**, 199–212 (2023).

37. W. Li, R. Ni, B. Tian, P. Lopes, Electrotactile Improves Thermal Referral. arXiv 2605.00240 [Preprint] (2026). https://doi.org/10.48550/arXiv.2605.00240.

38. P. Kourtesis, F. Argelaguet, S. Vizcay, M. Marchal, C. Pacchierotti, Electrotactile Feedback Applications for Hand and Arm Interactions: A Systematic Review, Meta-Analysis, and Future Directions. *IEEE Transactions on Haptics* **15**, 479–496 (2022).

39. S. Lee, S. Jang, Y. Cha, Soft wearable thermo+touch haptic interface for virtual reality. *iScience* **27**, 111303 (2024).

40. J.-H. Kim, A. Vázquez-Guardado, H. Luan, J.-T. Kim, D. S. Yang, H. Zhang, J.-K. Chang, S. Yoo, C. Park, Y. Wei, Z. Christiansen, S. Kim, R. Avila, J. U. Kim, Y. J. Lee, H.-S. Shin, M. Zhou, S. W. Jeon, J. M. Baek, Y. Lee, S. Y. Kim, J. Lim, M. Park, H. Jeong, S. M. Won, R. Chen, Y. Huang, Y. H. Jung, J.-Y. Yoo, J. A. Rogers, A wirelessly programmable, skin-integrated thermo-haptic stimulator system for virtual reality. *PNAS* **121**, e2404007121 (2024).

41. R. Chen, X. Mai, D. Chiaradia, A. Frisoli, D. Leonardis, A Soft Fabric-Based Thermal Haptic Device for Virtual Reality. *IEEE Access* **14**, 33651–33664 (2026).

42. J. S. Dunn, D. A. Mahns, S. S. Nagi, Why does a cooled object feel heavier? Psychophysical investigations into the Weber's Phenomenon. *BMC Neurosci.* **18**, 4 (2017).

43. B. G. Green, S. J. Lederman, J. C. Stevens, The effect of skin temperature on the perception of roughness. *Sens. Processes* **3**, 327–333 (1979).

44. M. Shibahara, K. Sato, "Illusion of Wet Sensation by Controlling Temperature and Softness of Dry Cloth" in *Haptics: Perception, Devices, Control, and Applications*, F. Bello, H. Kajimoto, Y. Visell, Eds. (Springer International Publishing, Cham, 2016), pp. 371–379.

45. Y. Liu, S. Nishikawa, Y. ah Seong, R. Niiyama, Y. Kuniyoshi, "ThermoCaress: A Wearable Haptic Device with Illusory Moving Thermal Stimulation" in *Proceedings of the 2021 CHI Conference on Human Factors in Computing Systems* (Association for Computing Machinery, New York, NY, USA, 2021; https://dl.acm.org/doi/10.1145/3411764.3445777)*CHI '21*, pp. 1–12.

46. H. Son, H. Wang, Y. Singhal, J. R. Kim, Upper Body Thermal Referral and Tactile Masking for Localized Feedback. *IEEE Transactions on Visualization and Computer Graphics* **29**, 2211–2219 (2023).

47. Y. Singhal, D. Honrales, H. Wang, J. R. Kim, "Thermal In Motion: Designing Thermal Flow Illusions with Tactile and Thermal Interaction" in *Proceedings of the 37th Annual ACM Symposium on User Interface Software and Technology* (Association for Computing Machinery, New York, NY, USA, 2024; https://dl.acm.org/doi/10.1145/3654777.3676460)*UIST '24*, pp. 1–13.

48. B. G. Green, Localization of thermal sensation: An illusion and synthetic heat. *Perception & Psychophysics* **22**, 331–337 (1977).

49. G. Gange, J. Knibbe, Optimising Automatic Calibration of Electric Muscle Stimulation. *Proceedings of the AAAI Conference on Artificial Intelligence* **35**, 3759–3767 (2021).

50. C. S. Deolindo, J. F. Ehmsen, A. S. Courtin, A. G. Mitchell, C. E. Kraenge, N. Nikolova, M. G. Allen, F. Fardo, Assessing individual sensitivity to the Thermal Grill Illusion: A two-dimensional adaptive psychophysical approach. *J. Pain* **27**, 104732 (2025).

51. D.-H. Kim, N. Lu, R. Ma, Y.-S. Kim, R.-H. Kim, S. Wang, J. Wu, S. M. Won, H. Tao, A. Islam, K. J. Yu, T. Kim, R. Chowdhury, M. Ying, L. Xu, M. Li, H.-J. Chung, H. Keum, M. McCormick, P. Liu, Y.-W. Zhang, F. G. Omenetto, Y. Huang, T. Coleman, J. A. Rogers, Epidermal Electronics. *Science* **333**, 838–843 (2011).

52. J. C. Lawler, M. J. Davis, E. C. Griffith, Electrical Characteristics of the Skin: The Impedance of the Surface Sheath and Deep Tissues. *J. Invest. Dermatol.* **34**, 301–308 (1960).

53. J.-P. Koo, S.-H. Kim, H.-J. An, O.-G. Moon, J.-H. Choi, Y.-D. Yun, J.-H. Park, K.-O. Min, Two-point discrimination of the upper extremities of healthy Koreans in their 20's. *J. Phys. Ther. Sci.* **28**, 870–874 (2016).

54. B. Simon, D. Stanke, M. Rohs, "Funnelectro: Electrotactile Funneling Illusion and Localization Performance on the Forearm" in *Proceedings of the 24th International Conference on Mobile and Ubiquitous Multimedia* (Association for Computing Machinery, New York, NY, USA, 2025; https://dl.acm.org/doi/10.1145/3771882.3771890)*MUM '25*, pp. 78–89.

55. A. Higashiyama, G. B. Rollman, Perceived locus and intensity of electrocutaneous stimulation. *IEEE Trans. Biomed. Eng.* **38**, 679–686 (1991).

56. M. Isaković, J. Malešević, M. Kostić, S. Došen, M. Štrbac, The Impact of Size and Position of Reference Electrode on the Localization of Biphasic Electrotactile Stimulation on the Fingertips. *IEEE Transactions on Haptics* **15**, 255–266 (2022).

57. K. Lim, J. Lee, S. Kim, M. Oh, C. S. Koh, H. Seo, Y.-M. Hong, W. G. Chung, J. Jang, J. A. Lim, H. H. Jung, J.-U. Park, Interference haptic stimulation and consistent quantitative tactility in transparent electrotactile screen with pressure-sensitive transistors. *Nat. Commun.* **15**, 7147 (2024).

58. T. E. Nichols, A. P. Holmes, Nonparametric permutation tests for functional neuroimaging: A primer with examples. *Human Brain Mapping* **15**, 1–25 (2002).

59. B. Phipson, G. K. Smyth, Permutation P-values should never be zero: calculating exact P-values when permutations are randomly drawn. *Stat. Appl. Genet. Mol. Biol.* **9**, Article39 (2010).

60. T. Jodai, L. A. Jones, M. Terao, H.-N. Ho, Perceiving Synchrony: Determining Thermal-Tactile Simultaneity Windows. *IEEE Transactions on Haptics* **17**, 850–859 (2024).

61. H.-N. Ho, K. Sato, S. Kuroki, J. Watanabe, T. Maeno, S. Nishida, Physical-Perceptual Correspondence for Dynamic Thermal Stimulation. *IEEE Transactions on Haptics* **10**, 84–93 (2017).

62. H. Kajimoto, Y. Kanno, S. Tachi, "Forehead Electro-tactile Display for Vision Substitution" in *EuroHaptics* (2006; https://www.semanticscholar.org/paper/Forehead-Electro-tactile-Display-for-Vision-Kajimoto/4d61efe8fd515c55fb6f8c3f5c5bf8d48cffc2b4).

63. M. Churyukanov, L. Plaghki, V. Legrain, A. Mouraux, Thermal Detection Thresholds of Aδ- and C-Fibre Afferents Activated by Brief CO2 Laser Pulses Applied onto the Human Hairy Skin. *PLOS ONE* **7**, e35817 (2012).

64. A. B. Vallbo, R. S. Johansson, Properties of cutaneous mechanoreceptors in the human hand related to touch sensation. *Hum. Neurobiol.* **3**, 3–14 (1984).

**Acknowledgments:** We thank R. Ni for assistance with the VR scene, K. Yin for supporting the device design and Prof. S. Sarah for providing the TurtleBot 4 robot.

**Funding:** W.L. acknowledge the support from Benjamin Ball Freud Merit Scholarship (University of Chicago). B.T. discloses support for the research of this work from the Air Force Office of Scientific Research (FA9550261B103), and the US Army Research Office (W911NF-24-1-0053).

**Author contributions:**

Conceptualization: W.L., P.L.,

Methodology: W.L., Z.X, E.M.A., X.N., P.L.,

Investigation: W.L., E.M.A, A.E.,

Visualization: W.L., J.Y.,

Funding acquisition: X.N., B.T., P.L.,

Project administration: X.N., B.T., P.L.,

Supervision: W.L., X.N., B.T., P.L.,

Writing – original draft: W.L., Z.X., P.L.,

Writing – review & editing: W.L., E.M.A, Z.X., X.N., B.T., P.L.

**Competing interests:** Authors declare that they have no competing interests.

**Data, code, and materials availability:** All data needed to evaluate the conclusions in the paper are present in the paper or the Supplementary Materials. The hardware design KiCad file, electronic components needed for the PCB, firmware for driving the PCB, figure raw data, thermal and tactile prediction code for this study have been deposited in the database https://github.com/humancomputerintegration/electrotactile-thermal-illusion. No new materials were generated in this study.